\documentclass[sigconf]{acmart}
\usepackage{lipsum}  
\usepackage{xspace}
\usepackage[LGR,T1]{fontenc}
\usepackage[utf8]{inputenc}
\newcommand{\textgreek}[1]{\begingroup\fontencoding{LGR}\selectfont#1\endgroup}

\newcommand{\bertbase}{BERT-BASE\xspace}
\newcommand{\bert}{BERT\xspace}
\newcommand{\roberta}{ROBERTA\xspace}
\newcommand{\camembert}{CAMEMBERT\xspace}
\newcommand{\finbert}{FinBERT\xspace}
\newcommand{\xlm}{XLM\xspace}
\newcommand{\xlmr}{XLM-R\xspace}
\newcommand{\dam}{DAM\xspace}
\newcommand{\greekbert}{GREEK-BERT\xspace}
\newcommand{\mbert}{M-BERT\xspace}
\newcommand{\mbertcased}{M-BERT-CASED\xspace}
\newcommand{\mbertuncased}{M-BERT-UNCASED\xspace}
\newcommand{\bilstmcnncrf}{BILSTM-CNN-CRF\xspace}

\newcommand{\rnn}{RNN\xspace}
\newcommand{\pos}{PoS\xspace}
\newcommand{\ner}{NER\xspace}
\newcommand{\nli}{NLI\xspace}
\newcommand{\xnli}{XNLI\xspace}
\newcommand{\mlm}{MLM\xspace}
\newcommand{\tlm}{TLM\xspace}
\newcommand{\nsp}{NSP\xspace}
\newcommand{\cls}{\texttt{[CLS]}\xspace}
\newcommand{\sep}{\texttt{[SEP]}\xspace}

\newcommand{\glue}{GLUE\xspace}

\newcommand{\nlp}{NLP\xspace}
\newcommand{\squad}{SQUAD\xspace}
\newcommand{\race}{RACE\xspace}
\newcommand{\bpe}{BPE\xspace}
\newcommand{\wordpiece}{WordPiece\xspace}

\AtBeginDocument{%
  \providecommand\BibTeX{{%
    \normalfont B\kern-0.5em{\scshape i\kern-0.25em b}\kern-0.8em\TeX}}}

\copyrightyear{2020}
\acmYear{2020}
\setcopyright{acmcopyright}\acmConference[SETN 2020]{11th Hellenic Conference on Artificial Intelligence}{September 2--4, 2020}{Athens, Greece}
\acmBooktitle{11th Hellenic Conference on Artificial Intelligence (SETN 2020), September 2--4, 2020, Athens, Greece}
\acmPrice{15.00}
\acmDOI{10.1145/3411408.3411440}
\acmISBN{978-1-4503-8878-8/20/09}

\begin{document}

\title{GREEK-BERT: The Greeks visiting Sesame Street}


\author{John Koutsikakis}
\authornotemark[1]
\author{Ilias Chalkidis}
\authornote{Equal contribution.}
\author{Prodromos Malakasiotis}
\author{Ion Androutsopoulos}
\email{[jkoutsikakis, ihalk, rulller, ion]@aueb.gr}
\affiliation{%
  \institution{Department of Informatics, Athens University of Economics and Business}
}

\renewcommand{\shortauthors}{Koutsikakis et al.}

\begin{abstract}
  Transformer-based language models, such as \bert and its variants, have achieved state-of-the-art performance in several downstream natural language processing (\nlp) tasks on generic benchmark datasets (e.g., \glue, \squad, \race). However, these models have mostly been applied to the resource-rich English language. In this paper, we present \greekbert, a monolingual \bert{-based} language model for modern Greek. We evaluate its performance in three \nlp tasks, i.e., part-of-speech tagging, named entity recognition, and natural language inference, obtaining state-of-the-art performance. Interestingly, in two of the benchmarks \greekbert outperforms two multilingual Transformer-based models (\mbert, \xlmr), as well as shallower neural baselines operating on pre-trained word embeddings, by a large margin (5\%-10\%). Most importantly, we make both \greekbert and our training code publicly available, along with code illustrating how \greekbert can be fine-tuned for downstream \nlp tasks. We expect these resources to boost \nlp research and applications for modern Greek.
\end{abstract}



\maketitle

\section{Introduction}
Natural Language Processing (\nlp) has entered its ImageNet  \cite{imagenet_cvpr09} era as advances in transfer learning have pushed the limits of the field in the last two years \cite{ruder2019}. Pre-trained language models based on Transformers~\cite{Vaswani2017}, such as \bert~\cite{devlin2019} and its variants \cite{roberta,xlnet,albert}, have achieved state-of-the-art results in several downstream \nlp tasks (e.g., text classification, natural language inference) on generic benchmark datasets, such as \glue~\cite{wang2018glue}, \squad~\cite{rajpurkar2016squad}, and \race~\cite{lai2017race}. However, these models have mostly targeted the English language, for which vast amounts of data are readily available. Recently, multilingual language models (e.g., \mbert, \xlm, \xlmr) have been proposed \cite{artetxe2018,lample2019, conneau2019} covering multiple languages, including modern Greek. While these models provide surprisingly good performance in zero-shot configurations (e.g., fine-tuning a pre-trained model in one language for a particular downstream task, and using the model in another language for the same task without further training), monolingual models, when available, still outperform them in most downstream tasks, with the exception of machine translation, where multilingualism is crucial. Consequently, Transformer-based language models have recently been adapted for, and applied to other languages \cite{martin2020camembert} or even specialized domains (e.g., biomedical \cite{alsentzer2019, beltagy2019}, finance \cite{araci2019finbert}, etc.) with promising results. To our knowledge, there are no publicly available pre-trained Transformer-based language models for modern Greek (hereafter simply Greek), for which much fewer \nlp resources are available, compared to English. Our main contributions are:

\begin{figure*}[!htbp]
    \centering
    \includegraphics[width=0.85\textwidth]{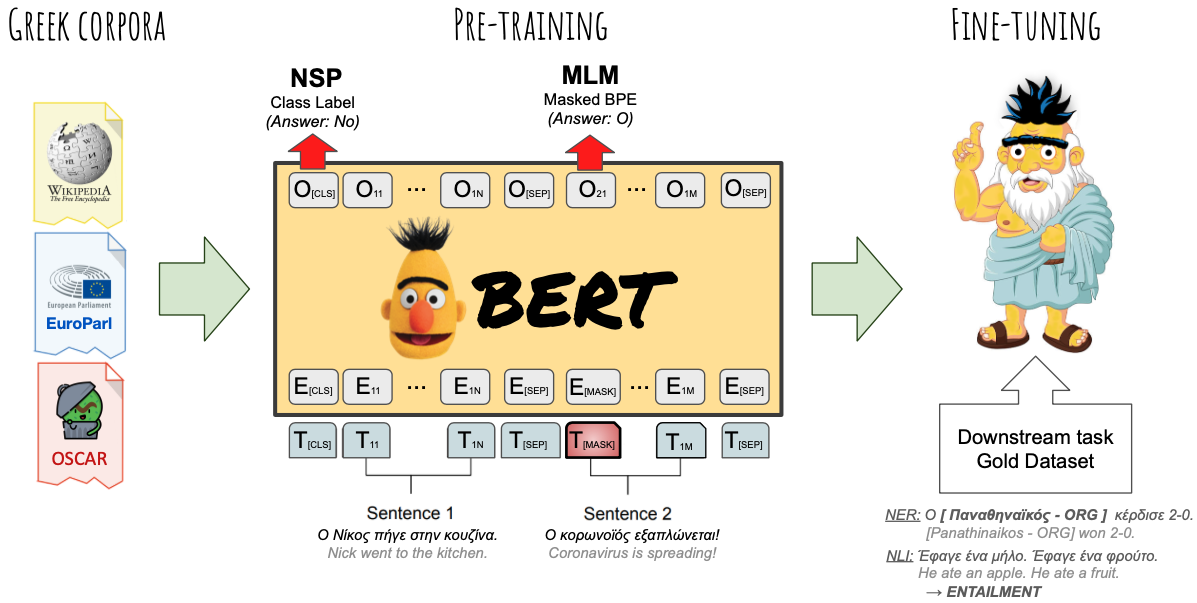}
    \caption{The two stages of employing \greekbert: (a) pre-traning \bert with the \mlm and \nsp objectives on Greek corpora, and (b) fine-tuning the pre-trained \bert model for downstream Greek \nlp tasks.}
    \label{fig:bert}
\end{figure*}

\begin{itemize}
    \item We introduce \greekbert, a new monolingual pre-trained Transformer-based language model for Greek, similar to \bertbase \cite{devlin2019}, trained on 29~GB of Greek text with a 35k sub-word 
    \bpe vocabulary created from scratch.
    \smallskip
    \item We compare \greekbert against multilingual language models based on Transformers 
    (\mbert, \xlmr) and other strong neural baselines operating on pre-trained word embeddings in three core downstream \nlp tasks, i.e., Part-of-Speech (\pos) tagging, Named Entity Recognition (\ner), and Natural Language Inference (\nli). \greekbert achieves state-of-the-art results in all datasets, while outperforming its competitors by a large margin (5-10\%) in two of them.
    \smallskip
    \item Most importantly, we make publicly available both the pre-trained 
    \greekbert and our training code, along with code illustrating how \greekbert can be fine-tuned for downstream \nlp tasks in Greek.\footnote{The pre-trained model is available at \url{https://huggingface.co/nlpaueb/bert-base-greek-uncased-v1}. The project is available at \url{https://github.com/nlpaueb/greek-bert}} We expect these resources to boost \nlp research and applications for Greek, since fine-tuning pre-trained Transformer-based language models for particular downstream tasks is the state-of-the-art. 
\end{itemize}

\section{Related Work}
\label{relatedwork}
\subsection{BERT: A Transformer language model}
\citet{devlin2019} introduced \bert, a deep language model based on Transformers \cite{Vaswani2017},  which is pre-trained on pairs of sentences to learn to produce high quality context-aware representations of sub-word tokens and entire sentences (Fig.~\ref{fig:bert}). Each sentence is represented as a sequence of \wordpiece{s}, a variation of \bpe{s} \cite{Gage1994}, 
while the special tokens \cls and \sep are used to indicate the start of the sequence and the end of a sentence, respectively. In other words, each pair of sentences has the form: $\left<\right.$\cls, \texttt{S-1}, \sep, \texttt{S-2}, \sep~$\left.\right>$, where \texttt{S1} is the first and \texttt{S2} the second sentence of the pair. \bert is pre-trained in two auxiliary self-supervised tasks: (a) \emph{Masked Language Modelling} (\mlm), also called \emph{Denoising Language Modelling} in the literature, where the model tries to predict masked-out (hidden) tokens based on the surrounding context, and (b) \emph{Next Sentence Prediction} (\nsp), where the model uses the representation of \cls to predict whether \texttt{S2} immediately follows \texttt{S1} or not, in the corpus they were taken from. The original English \bert was pre-trained on two generic corpora, English Wikipedia and Children's Books \cite{Zhu2015} with a vocabulary of 32k 
sub-words
extracted from the same corpora. \citet{devlin2019} originally released four English models. Two of them, BERT-BASE-UNCASED (12 layers of stacked Transformers, each of  768 hidden units, 12 attention heads, 110M parameters) and BERT-LARGE-UNCASED (24 layers, each of 1024 hidden units, 16 attention heads, 340M parameters), convert all text to lower-case. The other two models, BERT-BASE-CASED and BERT-LARGE-CASED, have the 
exact same architectures as the corresponding previous models, but preserve character casing.

\subsection{Multilingual language models}
Most work on transfer learning for languages other than English focuses on multilingual language modeling to cover multiple languages at once. Towards that direction, \mbert \cite{devlin2019}, a multilingual version of \bert, supports 100 languages, including Greek. \mbert was pre-trained with the same auxiliary tasks as \bert (\mlm, \nsp), on the Wikipedias of the supported languages. Each pre-training sentence pair contains sentences from the same language, but now pairs from all 100 languages are used. To cope with multiple languages, \mbert relies on an extended shared vocabulary of 110k 
sub-words. Only a small portion of the vocabulary (1,208 \wordpiece{s} or approx. 1\%) applies to the Greek language, mainly because of the Greek alphabet. By contrast, languages that use the Roman alphabet (e.g., English, French) share more sub-words \cite{conneau2018}, which as shown by \citet{lample2018unsupervised} greatly improves the alignment of embedding spaces across these languages. \mbert has been mainly evaluated as a baseline for zero-shot cross-lingual training \cite{lample2019}.

More recently, \citet{lample2019} introduced \xlm, a multilingual language model pre-trained on the Wikipedias of 15 languages, including Greek. They reported state-of-the-art results in supervised and unsupervised machine translation and cross-lingual classification. Similarly to \mbert, \xlm was trained in two auxiliary tasks, \mlm and the newly introduced \emph{Translation Language Modeling} task (\tlm). \tlm is a supervised extension of \mlm, where each training pair contains two sentences, \texttt{S1} and \texttt{S2}, from two different languages, \texttt{S1} being the translation of \texttt{S2}. When a word of \texttt{S1} (or \texttt{S2}) is masked, the corresponding translation of that word in \texttt{S2} (or \texttt{S1}) is not masked. In effect, the model learns to align representations across languages. \citet{conneau2019} introduced \xlmr, which further improved the results of \xlm, without relying on the supervised \tlm, by using more training data from Common Crawl, and a larger vocabulary of 250k sub-words, covering 100 languages. Again, a small portion of the vocabulary covers the Greek language (4,862 sub-words, approx. 2\%).

\subsection{Monolingual language models}
\citet{martin2020camembert} released \camembert, a monolingual language model for French, based on \roberta \cite{roberta}.  \camembert reported state-of-the-art results on four downstream tasks for French (\pos tagging, dependency parsing, \ner, \nli), outperforming \mbert and \xlm among other neural methods. \finbert \cite{virtanen2019} is another monolingual language model, for Finish, based on \bert. It achieved state-of-the-art results, in PoS tagging, dependency parsing, \ner, and text classification, outperforming \mbert among other models. Monolingual Transformer-based models have also been released for other languages (e.g., Italian, German, Spanish, etc.), showing strong performance in preliminary experiments. Most of them are still under development with no published work describing them.\footnote{An extensive list of monolingual Transformer-based models can be found at \url{https://huggingface.co/models}, along with preliminary results (when available).} 

\subsection{\nlp in Greek}
Publicly available resources for Greek \nlp continue to be very limited, compared to more widely spoken languages, although there have been several efforts to develop \nlp datasets, tools, and infrastructure for Greek \nlp.\footnote{Consult \url{https://www.clarin.gr/el/content/nlpel-clarin-knowledge-centre-natural-language-processing-greece} and \url{http://nlp.cs.aueb.gr/software.html} for examples.} Deep learning resources for Greek \nlp are even more limited. Recently, \citet{outsios2019} presented an evaluation of Greek word embeddings, comparing Greek Word2Vec \cite{mikolov2013} models against the publicly available Greek FastText \cite{bojanowski2017} model. Neural \nlp models that only pre-train word embeddings, however, have been largely superseded by deep pre-trained Transformer-based language models in the last two years. To the best of our knowledge, no Transformer-based pre-trained language model especially for Greek has been published to date. Thus, we aim to develop, study, and release such an important resource, as well as to provide an extensive evaluation across several \nlp tasks, comparing against state-of-the-art models. Part of our study compares against strong neural, in most cases \rnn{-based}, methods relying on pre-trained word embeddings. Such neural models are often not considered in other recent studies, without convincing justification. Instead, \emph{monolingual} pre-trained Transformer-based language models are solely compared to \emph{multilingual} pre-trained Transformer-based ones. We suspect that the latter, being usually biased towards more resource-rich languages (e.g., English, French, Spanish, etc.), may be outperformed by shallower neural models that rely only on pre-trained word embeddings. Hence, models of the latter kind may be stronger baselines for GREEK-BERT.

\section{GREEK-BERT}
\label{model}

In this work, we present \greekbert, a new monolingual version of \bert for Greek. We use the architecture of BERT-BASE-UNCASED, because the larger BERT-LARGE-UNCASED architecture is computationally heavy for both pre-training and fine-tuning. We pre-trained \greekbert on 29 GB of text from the following corpora: (a) the Greek part of Wikipedia;\footnote{An up-to-date dump can be found at \url{https://dumps.wikimedia.org/elwiki/}.} (b) the Greek part of the European Parliament Proceedings Parallel Corpus (Europarl) \cite{europarl}; and (c) the Greek part of OSCAR \cite{oscar}, a clean version of Common Crawl.\footnote{\url{https://commoncrawl.org}} Accents and other diacritics were removed, and all words were lower-cased to provide the widest possible normalization.
The same corpora were used to extract a vocabulary of 35k \bpe{s} with the SentencePiece library \cite{kudo-richardson2018}. Table~\ref{tab:datasets} presents the statistics of the pre-training corpora.
To pre-train \greekbert in the \mlm and \nsp tasks, we used the official code provided by Google.\footnote{\url{https://github.com/google-research/bert}} Similarly to \citet{devlin2019}, we used 1M pre-training steps and the Adam optimizer \cite{kingma14} with an initial learning rate of $1\mathrm{e}$-$4$ on a single Google Cloud TPU v3-8.\footnote{The Google Cloud TPU v3-8 was provided for free by the TensorFlow Research Cloud (TFRC) program, to which we are grateful.} Pre-training took approx.\ 5 days.

\begin{table}[!ht]
    \centering
    \begin{tabular}{l|ccc}
         Corpus & Size (GB) & Training pairs (M) & Tokens (B) \\
         \hline
         Wikipedia &  0.73 & 0.28  & 0.08 \\
         Europarl &  0.38 & 0.14  & 0.04 \\
         OSCAR &  27.0 & 10.26  & 2.92 \\
         \hline
         Total & 29.21 & 10.68 & 3.04 \\
    \end{tabular}
    \caption{Statistics on pre-training corpora for \greekbert.}
    \label{tab:datasets}
    \vspace{-3mm}
\end{table}

\section{Benchmarks}
\label{benchmarks}

We compare the performance of \greekbert against strong baselines on datasets for three core \nlp downstream tasks.

\subsection{Part-of-Speech tagging}

For the first downstream task, \pos tagging, we use the Greek Universal Dependencies Treebank (GUDT) \cite{prokopidis2005, prokopidis2017},\footnote{\url{https://github.com/UniversalDependencies/UD_Greek-GDT}.} which has been derived from the Greek Dependency Treebank,\footnote{\url{http://gdt.ilsp.gr}} a resource developed and maintained by the Institute for Language and Speech Processing, Research Center `Athena'.\footnote{\url{http://www.ilsp.gr/}} The dataset contains 2,521 sentences split in \emph{train} (1,622), \emph{development} (403), and \emph{test} (456) sets. The sentences have been annotated with \pos tags from a collection of 17 universal \pos tags (UPoS).\footnote{Additional information on the curation of the dataset can be found at \url{https://universaldependencies.org/treebanks/el_gdt/index.html}.} We ignore the syntactic dependencies of the dataset, since we consider only \pos tagging.

\subsection{Named Entity Recognition}

For the second downstream task, named entity recognition (\ner), we use two currently unpublished datasets, developed by I.~Darras and A.~Romanou, during student projects at NTUA and AUEB, respectively.\footnote{The annotated dataset of I.~Darras was part of his project for Google Summer of Code 2018 (\url{https://github.com/eellak/gsoc2018-spacy}), while A.~Romanou annotated documents with named entities for another project (\url{http://greekner.me/info}). We are grateful to both for sharing their datasets.} 
As both datasets, are fairly small, containing 1,798 and 2,521 sentences, we merged them and eliminated duplicate sentences. The merged dataset contains 4,189 unique sentences. 
We use the \emph{Person}, \emph{Organization}, and \emph{Location} annotations only, since the other entity types are not shared across the two datasets. 

\subsection{Natural Language Inference}

Finally, we experiment with the Cross-lingual Natural Language Inference corpus (\xnli) \cite{conneau2018}, which contains 5,000 test and 2,500 development pairs from the MultiNLI corpus \cite{adina2018}. Each pair consists of a \emph{premise} and a \emph{hypothesis}, and the task is to decide whether the premise entails (E), contradicts (C), or is neutral (N) to the hypothesis.
The test and development pairs, originally in English, were manually classified (as E, C, N) by crowd-workers, and they were then manually translated by professional translators (using the \emph{One Hour Translation} platform) to 14 languages, including Greek. The premises and hypotheses were translated separately, to ensure that no context is added to the hypothesis that was not there originally. Hence, each pair is available in 14 languages, always with the same class label.
MultiNLI also has a training set of 340k pairs. In \xnli \cite{conneau2018}, the training set of 340k pairs was automatically translated from English to the other languages, hence its quality is questionable; we discuss this further below. 
Although \xnli has been mainly used as a cross-lingual test-bed for multilingual models \cite{artetxe2018, lample2019}, we only consider its Greek part, i.e., we only use Greek pairs.

\section{Experimental Setup}
\label{experimentalsetup}

\subsection{Multilingual models}

For each task and dataset, we compare \greekbert against \xlmr and both the cased and uncased versions of \mbert.\footnote{The \bpe vocabulary of \xlmr retains both character casing and accents. We use the BASE version of \xlmr to be comparable with the rest of \bert models. The models are available at \url{https://huggingface.co/xlm-roberta-base}, \url{https://huggingface.co/bert-base-multilingual-cased}, and \url{https://huggingface.co/bert-base-multilingual-uncased}.} Recall that these models cover Greek with just a small portion of their sub-word vocabularies (approx. 2\% for \xlmr and 1\% for \mbert), which may cause excessive word fragmentation.

\begin{table}[!ht]
    \centering
    \begin{tabular}{l|c|c|c}
         Model & GUDT (\pos) & NER & XNLI \\
         \hline
         \mbertuncased \cite{devlin2019} & 2.38 & 2.43 & 2.22 \\
         \mbertcased \cite{devlin2019} & 2.58 & 2.65 & 2.40 \\
         \xlmr \cite{conneau2019} & 1.82 & 1.92 & 1.64\\
         \greekbert (ours) & \textbf{1.35} & \textbf{1.33} & \textbf{1.23} \\
    \end{tabular}
    \caption{Word fragmentation ratio of \bert{-based} models calculated on the development data of all datasets. Lower ratios are better. Best results shown in bold.}
    \label{tab:word_fragmentation}
    \vspace{-3mm}
\end{table}

Table~\ref{tab:word_fragmentation} reports the word fragmentation ratio, measured as the average number of sub-word tokens per word, in the three datasets for all Transformer-based language models considered. The multilingual models tend to fragment the words more than \greekbert, especially \mbert variants, whose fragmentation ratio is approximately twice as large. \xlmr has a lower fragmentation ratio than \mbert, as it has four times more sub-words covering Greek. 

All three  multilingual models often over-fragment Greek words. For example, in \mbertuncased `\textgreek{κατηγορουμενος}'  becomes
[`\textgreek{κ}', `\textgreek{\_ατ}', `\textgreek{\_η'}, `\textgreek{\_γο}', `\textgreek{\_ρου}', `\textgreek{\_μενος}'], and `\textgreek{γνωμη}' 
becomes [`\textgreek{γ}', `\textgreek{\_ν}', `\textgreek{\_ωμη}'], but both words exist in \greekbert's vocabulary. We suspect that, despite the ability of sub-words to effectively prevent out-of-vocabulary word instances, such long sequences of meaningless sub-words may be difficult to re-assemble into meaningful units, even when using deep pre-trained models. By contrast, baselines that operate on embeddings of entire words do not suffer from this problem, which is one more reason to compare against them.

\subsection{Baselines operating on word embeddings}
For both \pos tagging and \ner, we experiment with an established neural sequence tagging model, dubbed \bilstmcnncrf, introduced by \citet{ma-hovy-2016}. This model initially maps each word to the corresponding pre-trained embedding ($\boldsymbol{e}_i$), as well as to an embedding ($\boldsymbol{c}_i$) produced from the characters of the word by a Convolutional Neural Network (CNN). 
The two embeddings of each word are then concatenated  ($\boldsymbol{w}_i=[\boldsymbol{e}_i;\boldsymbol{c}_i]$). 
Each text $T$ is viewed as a sequence of embeddings $ \left<\boldsymbol{w}_1,\dots,\boldsymbol{w}_{|T|}\right>$. A stacked bidirectional LSTM \cite{hochreiter1997} turns the latter to a sequence of context-aware embeddings, which is fed to a linear Conditional Random Field (CRF) \cite{lafferty01} layer to produce the final predictions. 

For \nli, we re-implemented the Decomposable Attention Model (DAM) \cite{parikh2016}, which consists of an attention, a comparison, and an aggregation component. The attention component measures the importance of each word of the premise with respect to each hypothesis word and vice versa, as the normalized (via $\mathrm{softmax}$) similarity of all possible pairs of words between the hypothesis and the premise. The similarities are calculated as the dot products of the corresponding word embeddings, which are first projected through a shared MLP layer. Finally, each word of the premise (or hypothesis) is represented by an attended embedding ($\boldsymbol{a}_i$), which is simply the similarity weighted average of the embeddings of the hypothesis (or premise). The comparison component concatenates each $\boldsymbol{a}_i$ with the corresponding initial word embedding ($\boldsymbol{e}_i$) and projects the new representation through an MLP. In effect, the premise and the hypothesis are each represented by a set of comparison vectors
($P=\{\boldsymbol{v}_1,\dots,\boldsymbol{v}_p\}$, $H=\{\boldsymbol{u}_1,\dots,\boldsymbol{u}_h\}$, respectively). Finally, the aggregation component uses an MLP classifier for the final prediction, which operates on the concatenation of $\boldsymbol{s}_p$ with $\boldsymbol{s}_h$, where $\boldsymbol{s}_p=\sum_{\boldsymbol{v}_i\in P}\boldsymbol{v}_i$ and $\boldsymbol{s}_h=\sum_{\boldsymbol{u_i}\in H}\boldsymbol{u}_i$.

\subsection{Implementation details and hyper-parameter tuning}

All the baselines that require pre-trained word embeddings use the Greek FastText \cite{bojanowski2017} model\footnote{\url{https://fasttext.cc}} to obtain 300-dimensional word embeddings. The code for all experiments on downstream tasks is written in Python with the PyTorch\footnote{\url{https://pytorch.org}} framework using the PyTorch-Wrapper library \footnote{\url{https://github.com/jkoutsikakis/pytorch-wrapper}}. For the \bert-based models, we use the Transformers library from HuggingFace \cite{wolf2019huggingfaces}.\footnote{\url{https://github.com/huggingface/transformers}} The best architecture for each model is selected with grid search hyper-parameter tuning, minimizing the development loss, using early stopping without a maximum number of training epochs. For \bert models, we tuned the learning rate considering the range $\{2\mathrm{e}\text{-}5, 3\mathrm{e}\text{-}5, 5\mathrm{e}\text{-}5\}$, the dropout rate in the range $\{0, 0.1, 0.2\}$, and the batch size in $\{16, 32\}$. For \bilstmcnncrf, we used 2 stacked bidirectional LSTM layers and tuned the number of hidden units per layer in $\{100, 200, 300\}$, the learning rate in $\{1\mathrm{e}\text{-}2, 1\mathrm{e}\text{-}3\}$, the dropout rate in $\{0, 0.1, 0.2, 0.3\}$, and the batch size in $\{16, 32, 64\}$. Finally, for DAM we used 1 hidden layer with 200 hidden units for each MLP, and tuned the learning rate in $\{1\mathrm{e}\text{-}2, 1\mathrm{e}\text{-}3, 1\mathrm{e}\text{-}4\}$, the dropout rate in $\{0, 0.1, 0.2, 0.3\}$, and the batch size in $\{16, 32, 64\}$. Given the best hyper-parameter values, we train each model 3 times with different random seeds and report the mean scores and unbiased standard deviation (over the 3 repetitions) on the test set of each dataset.

\subsection{Denoising XNLI training data}

As already discussed, \xnli includes 2,500 development pairs, 5,000 test pairs, and 340k training pairs, which have been translated from English to 14 languages, including Greek. The training pairs were machine-translated and, unfortunately, many of the resulting training pairs are of very low quality, which may harm performance. Based on this observation, we wanted to assess the effect of using the full training set, including many noisy pairs, against using a subset of the training set containing only high-quality pairs. We estimate the quality of a machine-translated pair as the perplexity of the concatenated sentences of the pair, computed by using \greekbert as a language model, masking one \bpe of the two concatenated sentences at a time.\footnote{See Appendix A for a selection of random noisy samples and the best and worst samples according to \greekbert.} We retain the 40k training pairs (approx. 10\%) with the lowest (best) perplexity scores as the high-quality \xnli training subset.
For comparison, we also train (fine-tune) our models on the full \xnli training set. Unlike all other experiments, when using the full \xnli training set we do not perform three iterations (with different random seeds), because of the size of the full training set, which makes repeating experiments computationally much more expensive.

\section{Experimental Results}
\label{experiments}

Table~\ref{tab:pos_results} reports the \pos tagging results. All Transformer-based models have comparable performance (97.8-98.2 accuracy), and \xlmr is marginally (+0.1\%) better than \greekbert and \mbertcased. By contrast, \bilstmcnncrf performs clearly worse, but the difference from the other models is small (0.8-1.2\%).
The fact that all models achieve high scores can be explained by the observation that the correct \pos tag of a word in Greek can often be determined by considering mostly the word's suffix, or for short function words (e.g., determiners, prepositions) the word itself, and to a lesser extent the word's context. Thus, even multi-lingual models with a high word fragmentation ratio (\mbert, \xlmr) are often able to guess the \pos tags of words from their sub-words, even if the sub-words correspond to suffixes, other small parts of words, or frequent short words included in the sub-word vocabulary. Hence, there is often no need to consider the context of each word. The latter is difficult if context information gets scattered across too many very short sub-words.\footnote{In all multilingual models, each word is usually split in 2 or more sub-words and the last one resembles a suffix (e.g., `\textgreek{ανησυχιες}' becomes [`\textgreek{ανησυχ}', `\textgreek{\_ιες}'], `\textgreek{χαρακτηριζεται} becomes [`\textgreek{χαρακτηρ}', `\textgreek{\_ιζεται}']
), which often suffices to identify \pos tags.} The \bilstmcnncrf method, which uses word embeddings, is also able to exploit information from suffixes, because it produces extra word embeddings from the characters of the words (using a CNN).

\begin{table}[!ht]
    \centering
    \begin{tabular}{l|l}
    Model & Accuracy  \\
    \hline
    \bilstmcnncrf \cite{ma-hovy-2016} & 97.0 $\pm$ 0.14 \\
    \mbertuncased \cite{devlin2019}  & 97.8 $\pm$ 0.03 \\
    \mbertcased \cite{devlin2019} & 98.1  $\pm$ 0.08 \\
    \xlmr \cite{conneau2019} & \textbf{98.2}  $\pm$ 0.07 \\
    \greekbert (ours) & 98.1 $\pm$ 0.08 \\
    \end{tabular}
    \caption{\pos tagging results ($\pm$ std) on test data.}
    \label{tab:pos_results}
    \vspace{-3mm}
\end{table}

For a more complete comparison we also report results per \pos tag for the two best models, i.e., \greekbert and \xlmr (Table~\ref{tab:pos_results_full}). We again observe that the two models have almost identical performance. Interestingly, both models have difficulties in predicting the tag \emph{other} (X) and to a lesser extent \emph{proper nouns} (PROPN) and \emph{numerals} (NUM). In fact, both models tend to confuse these \pos tags, which is reasonable considering that the inflectional morphology and context of Greek proper nouns is similar to that of common nouns, numerals (when written as words, e.g., `\textgreek{χιλιαδες}') often have similar morphology with nouns, and words tagged as `other' (X) are often foreign proper nouns; for example, `\textgreek{καστρο}' could either refer to `Fidel Castro' (X) or be the Greek noun for `castle'. Overall, there is still room for improvement in these particular tags.

\begin{table}[!ht]
    \centering
    \begin{tabular}{l|c|c}
         Part-of-Speech tag &  \greekbert & \xlmr \\
         \hline
        ADJ   &  95.6 $\pm$ 0.26 &  96.0 $\pm$ 0.17 \\
        ADP   &  99.7 $\pm$ 0.07 &  99.8 $\pm$ 0.03 \\
        ADV   &  97.2 $\pm$ 0.34 &  97.4 $\pm$ 0.12 \\
        AUX   &  99.9 $\pm$ 0.15 &  99.8 $\pm$ 0.20 \\
        CCONJ &  99.6 $\pm$ 0.24 &  99.7 $\pm$ 0.14 \\
        DET   &  99.8 $\pm$ 0.08 &  99.9 $\pm$ 0.02 \\
        NOUN  &  97.9 $\pm$ 0.28 &  97.9 $\pm$ 0.08 \\
        NUM   &  92.7 $\pm$ 1.14 &  93.0 $\pm$ 0.85 \\
        PART  & 100.0 $\pm$ 0.00 &  99.7 $\pm$ 0.45 \\
        PRON  &  98.8 $\pm$ 0.25 &  98.6 $\pm$ 0.21 \\
        PROPN &  86.0 $\pm$ 1.03 &  87.0 $\pm$ 0.37 \\
        PUNCT & 100.0 $\pm$ 0.00 & 100.0 $\pm$ 0.03 \\
        SCONJ &  99.4 $\pm$ 0.56 &  99.5 $\pm$ 0.16 \\
        VERB  &  99.3 $\pm$ 0.13 &  99.4 $\pm$ 0.16 \\
        X     &  77.3 $\pm$ 1.32 &  77.4 $\pm$ 2.16 \\
    \end{tabular}
    \caption{F1 scores ($\pm$ std) per \pos tag for the two best models (\greekbert, \xlmr). We do not report results for symbols (SYM) as there are no such annotations in the test data.}
    \label{tab:pos_results_full}
\end{table}

Table~\ref{tab:ner_results} presents the \ner results. We observe that \greekbert outperforms the rest of the methods, by a large margin in most cases; it is 9.3\% better than \bilstmcnncrf, 3.9\% better than the two \mbert models on average, and 0.9\% better than \xlmr. The \ner task is clearly more difficult than \pos tagging, as evidenced by the near-perfect performance of all methods in \pos tagging (Table~\ref{tab:pos_results}), compared to the far from perfect performance of all methods in \ner (Table~\ref{tab:ner_results}). Being more difficult, the \ner task leaves more space for better methods to distinguish themselves from weaker methods, and indeed \greekbert clearly performs better than all the other methods, with \xlmr being the second best method.

\begin{table}[!ht]
    \centering
    \begin{tabular}{l|c}
    Model & Micro-F1\\
    \hline
    \bilstmcnncrf \cite{ma-hovy-2016} & 76.4 $\pm$ 2.07\\
    \mbertuncased \cite{devlin2019}  & 81.5 $\pm$ 1.77\\
    \mbertcased \cite{devlin2019} & 82.1  $\pm$ 1.35\\
    \xlmr \cite{conneau2019} & 84.8  $\pm$ 1.50\\
    \greekbert (ours) & \textbf{85.7} $\pm$ 1.00\\
    \end{tabular}
    \caption{\ner results ($\pm$ std) on test data.}
    \label{tab:ner_results}
    \vspace{-3mm}
\end{table}

In Table~\ref{tab:per_entity}, we conduct a per entity type evaluation of the two best \ner models. Both models are more accurate when predicting \emph{persons} and \emph{locations}. \greekbert is better on persons, and \xlmr marginally better on locations. Concerning \emph{organizations}, \greekbert is better, but both models struggle, because organizations often contain person names (`\textgreek{μπισκότα \emph{Παπαδοπούλου}}') or locations (`\textgreek{Αθλέτικο \emph{Μπιλμπάο}}'), shown in italics.

\begin{table}[!ht]
    \centering
    \begin{tabular}{l|c|c}
         Entity type &  \greekbert & \xlmr \\
         \hline
         PERSON & 88.8 $\pm$ 3.06 & 85.2 $\pm$ 1.25 \\
         LOCATION & 88.4 $\pm$ 0.88 & 88.5 $\pm$ 0.86 \\
         ORGANIZATION & 69.6 $\pm$ 4.28 & 68.9 $\pm$ 5.62 \\
    \end{tabular}
    \caption{F1 scores ($\pm$ std) per entity type for the two best models (\greekbert, \xlmr) on test data.}
    \label{tab:per_entity}
\end{table}

Finally, Table~\ref{tab:xnli_results} shows that \greekbert is again substantially better than the rest of the methods in \nli, outperforming \dam (+10.1\%), the two \mbert models (+4.9\% on average) and \xlmr (+1.3\%). Interestingly, performance improves for all models when trained on the entire training set, as opposed to training only on the high-quality 10\% training subset, contradicting our assumption that noisy data could harm performance. Using a larger training set seems to be better than using a smaller one, even if the larger training set contains more noise. We suspect that noise may, in effect, be acting as a regularizer, improving the generalization ability of the models. A careful error analysis would shed more light on this phenomenon, but we leave this investigation for future work.

\begin{table}[!ht]
    \centering
    \resizebox{\columnwidth}{!}{
    \begin{tabular}{l|c|c}
    Training Data & 10\% high quality &  all train data \\
    \hline
    Model & Accuracy & Accuracy  \\
    \hline
    \dam \cite{parikh2016}           & 61.5 $\pm$ 0.94          & 68.5 $\pm$ 1.71 \\
    \mbertuncased \cite{devlin2019}  & 65.7 $\pm$ 1.01          & 73.9 $\pm$ 0.64 \\
    \mbertcased \cite{devlin2019}    & 64.6 $\pm$ 1.29          & 73.5 $\pm$ 0.49 \\
    \xlmr \cite{conneau2019}         & 70.5 $\pm$ 0.69          & 77.3 $\pm$ 0.41 \\
    \greekbert (ours)                & \textbf{71.6} $\pm$ 0.80  & \textbf{78.6} $\pm$ 0.62 \\
    \end{tabular}
    }
    \caption{\nli results ($\pm$ std) on test data.}
    \label{tab:xnli_results}
    \vspace{-3mm}
\end{table}

As in the previous datasets, we report results per class for the two best models in Table~\ref{tab:per_class_nli}. \greekbert is better in all three classes, while both models have difficulties when predicting \emph{neutral} pairs, which tend to be confused with pairs containing \emph{contradiction}. 

\begin{table}[!ht]
    \centering
    \begin{tabular}{l|c|c}
         Class &  \greekbert & \xlmr \\
         \hline
         ENTAILMENT    & 78.8 $\pm$ 1.20 & 78.0 $\pm$ 0.70 \\
         CONTRADICTION & 81.2 $\pm$ 0.15 & 79.7 $\pm$ 0.53 \\
         NEUTRAL       & 75.9 $\pm$ 0.74 & 74.1 $\pm$ 0.50 \\
    \end{tabular}
    \caption{F1 scores ($\pm$ std) per \nli class for the two best models (\greekbert, \xlmr) on test data. Both models were trained on all training data.}
    \label{tab:per_class_nli}
\end{table}

\section{Conclusions and Future Work}
\label{conclusions}

We presented \greekbert, a new monolingual \bert{-based} language model for modern Greek, which has been pre-trained on large modern Greek corpora and can be fine-tuned (further trained) for particular \nlp tasks. The new model achieves state-of-the-art performance in Greek \pos tagging, named entity recognition, and natural language inference, outperforming strong baselines in the latter two, more difficult tasks. The baselines we considered included deep multilingual Transformer-based language models (\mbert, \xlmr), and shallower established neural methods (\bilstmcnncrf, DAM) operating on word embeddings. Most importantly, we release the pre-trained \greekbert model, code to replicate our experiments, and code illustrating how \greekbert can be fine-tuned for \nlp tasks. We expect that these resources will boost \nlp research and applications for Greek, a language for which public \nlp resources, especially for deep learning, are still scarce.

In future work, we plan to pre-train another version of \greekbert using even larger corpora. We plan to use the entire corpus of Greek legislation \cite{chalkidis2017}, as published by the National Publication Office,\footnote{Available at \url{http://www.et.gr}.} and the entire Greek corpus of EU legislation, as published in Eur-Lex.\footnote{EU legislation is translated in the 24 official EU languages.} Both corpora include well-written Greek text describing policies across many different domains (e.g., economy, health, education, agriculture). Following \cite{raffel2019}, who showed the importance of cleaning data when pre-training language models, we plan to discard noisy parts from all corpora, e.g., by filtering out documents containing tables or other non-natural text. We also plan to investigate the performance of \greekbert in more downstream tasks, including dependency parsing, to the extent that more Greek datasets for downstream tasks will become publicly available. It would also be interesting to pre-train \bert-based models for earlier forms of Greek, especially classical Greek, for which large datasets are available.\footnote{For example, \url{http://stephanus.tlg.uci.edu/}, \url{https://www.perseus.tufts.edu/hopper/}.} This could potentially lead to improved \nlp tools for classical studies.

\begin{acks}
This project was supported by the TensorFlow Research Cloud (TFRC) program that provided a Google Cloud TPU v3-8 for free, while we also used free Google Cloud Compute (GCP) research credits. We are grateful to both Google programs.
\end{acks}

\bibliographystyle{ACM-Reference-Format}
\bibliography{greekbert}

\appendix

\section{Examples from the Greek part of XNLI corpus}

\begin{table*}[!h]
        {\footnotesize
        \centering
        \resizebox{\textwidth}{!}{
    \begin{tabular}{p{0.4\textwidth}|p{0.4\textwidth}|c}
         \hline
         \multicolumn{3}{c}{\small{\textsc{Random noisy samples}}} \\
         \hline
         \textbf{Premise} & \textbf{Hypothesis} & \textbf{Label} \\
        \hline
         \textgreek{Η εννοιολογικά κρέμα κρέμα έχει δύο βασικές διαστάσεις - προϊόν και γεωγραφία.} &  \textgreek{Το προϊόν και η γεωγραφία είναι αυτά που κάνουν την κρέμα να κλέβει.} & neutral \\ 
         \hline
         \textgreek{Ένας από τους αριθμούς μας θα μεταφέρει τις οδηγίες σας λεπτομερώς.} &  \textgreek{Ένα μέλος της ομάδας μου θα εκτελέσει τις διαταγές σας με τεράστια ακρίβεια.} & entailment \\ 
         \hline
         \textgreek{Γκέι και λεσβίες.} &  \textgreek{Ετεροφυλόφιλους.} & contradiction \\ 
         \hline
         \textgreek{Η ταχυδρομική υπηρεσία ήταν η μείωση της συχνότητας παράδοσης.} &  \textgreek{Η ταχυδρομική υπηρεσία θα μπορούσε να είναι λιγότερο συχνή.} & entailment \\ 
         \hline
         \textgreek{Αυτή η ανάλυση συγκεντρωτική εκτιμήσεις από τις δύο αυτές μελέτες για την ανάπτυξη μιας λειτουργίας} c-R \textgreek{που συνδέει το} pm \textgreek{με τη χρόνια βρογχίτιδα.} &  \textgreek{Η ανάλυση αποδεικνύει ότι δεν υπάρχει σύνδεση μεταξύ} pm \textgreek{και βρογχίτιδα.} & contradiction \\ 
         \hline
         \multicolumn{3}{c}{\small{\textsc{Best samples according to \greekbert}}} \\
         \hline
         \textbf{Premise} & \textbf{Hypothesis} & \textbf{Label} \\
        \hline
        \textgreek{Τηγανητό κοτόπουλο, τηγανητό κοτόπουλο, τηγανητό κοτόπουλο.} & \textgreek{Χάμπουργκερ, χάμπουργκερ, χάμπουργκερ.} & contradiction \\
        \hline
         \textgreek{Τα τελευταία χρόνια, το κογκρέσο έχει λάβει μέτρα για να αλλάξει ριζικά τον τρόπο με τον οποίο οι ομοσπονδιακές υπηρεσίες κάνουν τη δουλειά τους.} &  \textgreek{Το Κογκρέσο έχει λάβει μέτρα για να αλλάξει ριζικά τον τρόπο με τον οποίο οι ομοσπονδιακές υπηρεσίες κάνουν τη δουλειά τους τα τελευταία χρόνια.} & entailment \\ 
         \hline
         \textgreek{Για παράδειγμα, ορισμένες ηλικιακές ομάδες φαίνεται να είναι πιο ευαίσθητες στην ατμοσφαιρική ρύπανση από άλλες.} &  \textgreek{Η ατμοσφαιρική ρύπανση δεν μπορεί να επηρεάσει όλες τις ηλικιακές ομάδες.} & contradiction \\ 
         \hline
         \textgreek{Οι επισκέπτες μπορούν να δουν τα δελφίνια να εκπαιδεύονται και να τρέφονται κάθε δύο ώρες από τις 10:00 το πρωί έως τις 4:00μ.μ.} &  \textgreek{Μπορείτε επίσης να δείτε τα δελφίνια να καθαρίζονται στις 6:00μ.μ.} & neutral \\ 
         \hline
         \textgreek{Ναι, δεν ξέρω, όπως είπα, πιστεύω ότι πιστεύω στην θανατική ποινή, αλλά αν καθόμουν στους ενόρκους και έπρεπε να πάρω την απόφαση, δεν θα ήθελα να είμαι αυτός που θα τα καταφέρει.} & \textgreek{Πιστεύω στην θανατική ποινή, αλλά δεν θα ήθελα να είμαι σε θέση να κάνω αυτή την επιλογή.} & entailment\\
         \hline
         \multicolumn{3}{c}{\small{\textsc{Worst samples according to \greekbert}}} \\
         \hline
         \textbf{Premise} & \textbf{Hypothesis} & \textbf{Label} \\
         \hline
         \textgreek{Τα μάτια του θολή με δάκρυα και αυτός σίδερο στο πίσω μέρος του λαιμού του.} &  \textgreek{Έκλαιγε.} & contradiction \\ 
         \hline
         Um-\textgreek{Βουητό πιο εσωτερική.} &  \textgreek{Συνηθισμένη} & contradiction \\ 
         \hline
         Bonifacio &  \textgreek{Επισκεφτείτε το} bonifacio \textgreek{δωρεάν.} & neutral \\ 
         \hline
         \textgreek{Μπόστον σέλτικς δεξιά} &  \textgreek{Ιντιάνα πέισερς, όχι};& entailment \\ 
          \hline
         \textgreek{Για μένα ο} juneteenth \textgreek{ανακάλεσε ειδικά τον αβεσσαλώμ, τον αβεσσαλώμ!} &  Juneteenth \textgreek{ειδικά υπενθύμισε αβεσσαλώμ} & entailment \\ 
         \hline
    \end{tabular}
    }}
    \caption{Examples of pairs in the Greek part of the XNLI corpus.}
    \label{tab:xnli_samples}
\end{table*}

Table~\ref{tab:xnli_samples} presents examples from the Greek part of XNLI including: (a) random noisy samples with morphology and syntax errors, (b) a selection from the best samples according to \greekbert language modeling perplexity, and (c) a selection from the worst samples according to \greekbert language modeling perplexity.

\end{document}